\documentclass[11pt,a4paper]{article}
\usepackage{times,latexsym}
\usepackage{url}
\usepackage[T1]{fontenc}
\usepackage{mathtools}
\usepackage{amsmath}
\usepackage{amsfonts}
\usepackage{microtype}
\usepackage{amssymb}
\usepackage{xspace}
\usepackage{hhline}
\usepackage{tabularx}
\usepackage{arydshln}
\usepackage{multirow}
\usepackage{adjustbox,booktabs}
\usepackage{mathtools}
\usepackage{makecell}

\usepackage[multiple]{footmisc}
\usepackage{footnote}

\usepackage[]{ACL2023}

\usepackage{xspace,mfirstuc,tabulary}

\usepackage{xspace}

\newcommand{\secref}[2][]{Section#1~\ref{#2}\xspace}
\newcommand{\figref}[2][]{Figure#1~\ref{#2}\xspace}
\newcommand{\tabref}[2][]{Table#1~\ref{#2}\xspace}
\newcommand{\eqnref}[2][]{Eqn#1.~(\ref{#2})\xspace}
\newcommand{\round}[1]{\ensuremath{\lfloor#1\rceil}}

\newcommand{\ex}[1]{\textit{#1}\xspace}
\newcommand{\gl}[1]{``{#1}''\xspace}
\newcommand{\z}{\phantom{0}}

\newcommand{\minPts}{\ensuremath{\text{minPts}}}

\title{Unsupervised Paraphrasing of Multiword Expressions}

\author{Takashi Wada$^{1,3}\thanks{~~This work was partially done when the first author was at Riken.}$ 
\,  Yuji Matsumoto$^{2}$
\, Timothy Baldwin$^{1,3}$
\,  Jey Han Lau$^{1}$ \\
    $^1$ School of Computing and Information Systems, The University of Melbourne \\
    $^2$ RIKEN Center for Advanced Intelligence Project (AIP) \\
    $^3$ Department of Natural Language Processing, MBZUAI \\[0ex]
    \texttt{twada@student.unimelb.edu.au}, \enspace \texttt{tb@ldwin.net}\\\texttt{yuji.matsumoto@riken.jp},\enspace \texttt{jeyhan.lau@gmail.com}
    }    

\date{}

\begin{document}
\maketitle
\begin{abstract}
We propose an unsupervised approach to paraphrasing multiword expressions (MWEs) in context. Our model employs only monolingual corpus data and pre-trained language models (without fine-tuning), and does not make use of any external resources such as dictionaries. We evaluate our method on the SemEval 2022 idiomatic semantic text similarity task, and show that it outperforms all unsupervised systems and rivals supervised systems.\footnote{Code is available at: \url{https://github.com/twadada/mwe-paraphrase}.}
\end{abstract}

\section{Introduction}

Multiword expressions (MWEs) are notoriously difficult to model because the meaning of the whole can diverge substantially from that of the component words, e.g.\ the meaning of \ex{swan~song} (\gl{final performance}) is far removed from its component words\footnote{The MWE is said to originate from an ancient legend that a swan sings beautifully before it dies.} \cite{Pain_in_the_neck,mwe}. This hampers the capacity of pre-trained language models such as BERT \cite{bert} to capture MWE semantics \cite{tayyar-madabushi-etal-2021-astitchinlanguagemodels-dataset, bart_idiom}. Similarly, non-native speakers tend to have difficulty understanding MWEs, especially those that have no equivalent in their native language  \cite{Irujo,arnaud_savignon_1996}.

In this work, we propose a method for paraphrasing non-literal MWEs (e.g.\ \ex{swan~song}) into more literal expressions (e.g.\ \ex{final~performance}) to aid both humans and machines to understand their meanings. Importantly, we perform this in a {\it fully unsupervised} way: our method uses only monolingual corpus data and off-the-shelf pre-trained masked language models (MLMs), and does not make use of any labelled data or lexical resources such as WordNet \cite{wordnet}, in contrast with previous work \cite{liu-hwa-2016-phrasal, zhou-etal-2021-pie, Zhou_Zeng_Gong_Bhat_2022}. We base our experiments on SemEval 2022 Task 2 \cite{tayyar-madabushi-etal-2022-semeval}, a task designed to evaluate how well models understand the semantics of MWEs in two high-resource languages (English and Portuguese) and one low-resource language (Galician). We show that our model generates high-quality paraphrases of MWEs and aids pre-trained models to produce better representations for sentences that contain idiomatic expressions. Compared to system submissions to the shared task, our method performs better than all unsupervised systems and comparably with supervised systems.

\section{Methodology}
\begin{figure*}[!t]
\begin{center}
\includegraphics[bb=000 015 550 130,scale=1.0]{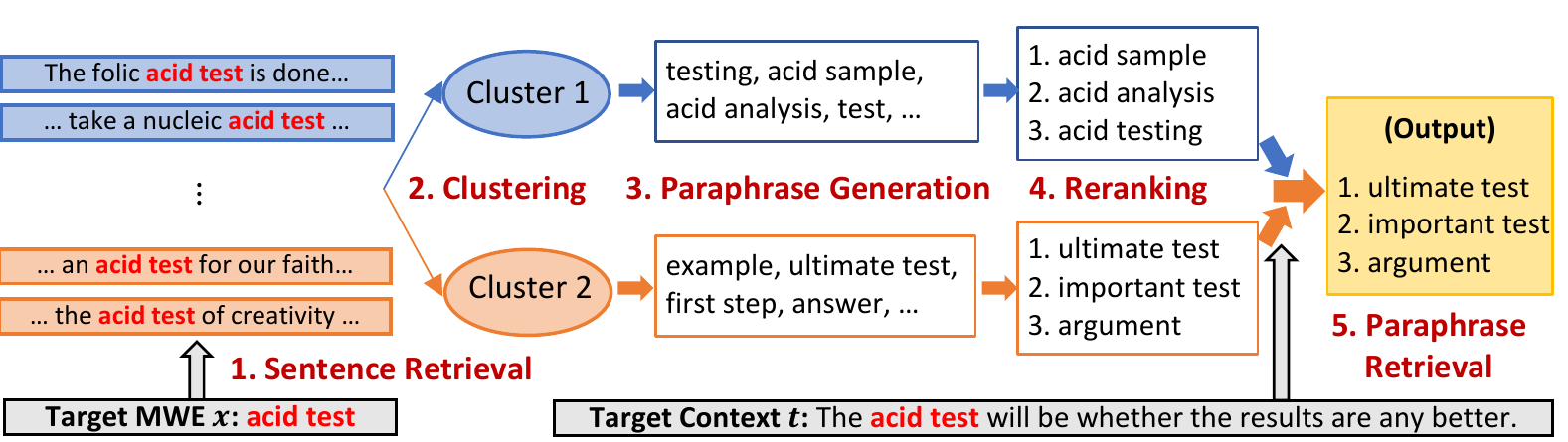}
\end{center}
\vspace*{5mm}
\caption{Overview of our proposed method.}
\label{model}
\end{figure*}
Given a target sentence $t$ that contains an MWE $x$, our goal is to paraphrase $x$ with more literal expressions in the context of $t$. To this end, we propose a fully unsupervised method which employs monolingual corpus data and an off-the-shelf masked language model as is, i.e.\ without any fine-tuning. Our method consists of the following five steps: (1) collect sentences containing $x$ from a monolingual corpus; (2) cluster the sentences; (3) generate paraphrases of $x$ for each cluster; (4) rerank the paraphrase candidates; and (5) select the most relevant cluster to the sense of $x$ in the target sentence $t$, and paraphrase $x$.
\figref{model} illustrates the overview of our model, and we describe the details of each step below.

\subsection{Sentence Retrieval and Clustering} \label{step1}
We first collect sentences that contain the target MWE $x$ from a monolingual corpus.\footnote{Here, we do not perform lemmatisation of $x$ and regard, say, \ex{ghost town} and \ex{ghost towns} as different instances because we aim to generate paraphrases that fit well both syntactically and semantically in the target sentence $t$.} Next, we sparsify sentences that have very similar local contexts around $x$ to ensure diversity in the data,\footnote{We look at 3 words surrounding $x$ for each sentence, and discard sentences that share many words with other sentences.} and keep up to 300 sentences $\{s_i\}$ for each MWE type. Then, we generate contextualised embeddings of $x$ for each sentence and cluster them to try to separate out different senses of $x$. For instance, for $x = $ \ex{closed book}, there should be at least two clusters to represent its literal sense (\gl{unopened book}) and idiomatic sense (\gl{mystery}; e.g.\ \ex{This subject is a \underline{closed book} to me}). To generate the embeddings, we replace $x$ with a single [MASK] token and obtain the representation of [MASK] just before the linear layer prior to the word prediction output.

As our clustering algorithm, we choose DBSCAN \cite{dbscan} since it adaptively determines the number of clusters (expected to roughly correspond to the number of senses of $x$), and dynamically removes outliers during clustering. We measure the distance using cosine similarity, and tune a couple of hyperparameters of DBSCAN on the dev set of the SemEval STS task.\footnote{The tuned values are shown in Appendix \ref{hyper_param}.} In \secref{effects_clustering_sec}, we show that DBSCAN is more effective than two other popular clustering methods: $K$-means \cite{kmeans,kmeans_pp} and X-means \cite{xmeans}.  As monolingual corpora, we use OSCAR \cite{OSCAR} for English and Portuguese, and CC-100 \cite{wenzek-etal-2020-ccnet,conneau-etal-2020-unsupervised} for Galician. 

\subsection{Paraphrase Generation} \label{step3}

Given the $N_j$ sentences in the $j$-th cluster $S_j$: $s_{1,j}, s_{2,j} .., s_{N_j,j}$, we generate paraphrase candidates of $x$. The key idea is to find words or phrases that are a good fit for the contexts of $x$ in all sentences in $S_j$. To this end, we use BERT to generate 1- and 2-token paraphrases, independently. To predict 1-token candidates $y$, we simply replace $x$ in $s_{i,j}$ with a single [MASK] token (denoted as $\mathrm{M}$), and obtain the probability as:
\begin{gather}
\mathrm{P_1}(y|S_j) = g(\frac{1}{N_j}\sum_{i = 1}^{N_j}f(s_{i,j}^{x\rightarrow \mathrm{M}})_{\mathrm{M}}), \label{bert_1mask}
\end{gather}
where $s_{i,j}^{x\rightarrow \mathrm{M}}$ denotes the sentence $s_{i,j}$ with $x$ replaced by [MASK]; $f( s_{i,j}^{x\rightarrow \mathrm{M}})_{\mathrm{M}} \in \mathbb{R}^d$ denotes the [MASK] representation before the last linear layer; and $g$ denotes the last linear layer followed by the softmax function. We calculate the probability distribution after averaging the [MASK] embeddings over the $N_j$ sentences, and retrieve the top-10 words as the 1-token paraphrase candidates. 

Similarly, when we generate 2-token paraphrases $y$: $y_1y_2$, we replace $x$ with two [MASK] tokens (denoted as $\mathrm{M_1}\mathrm{M_2}$) and obtain the probability distribution for $y_k$ as:
\begin{gather}
\tilde{\mathrm{P}}(y_k|S_j) = g(\frac{1}{N_j}\sum_{i = 1}^{N_j}f(s_{i,j}^{x\rightarrow \mathrm{M_1M_2}})_{\mathrm{M_k}})\label{2_mask_eqn}.
\end{gather}
Since extracting the most probable words independently from $\tilde{\mathrm{P}}(y_1|S_j)$ and $\tilde{\mathrm{P}}(y_2|S_j)$ does not always result in a valid phrase, we first extract the top-5 words in $\tilde{\mathrm{P}}(y_1|S_j)$ and use each of them as the basis of  $\mathrm{M_1}$, and then compute the probability of filling $\mathrm{M_2}$ given $y_{1}$, as:
\begin{equation}
\mathrm{\tilde{P}}(y_2|y_1,S_j) = g(\frac{1}{N_j}\sum_{i = 1}^{N_j}f(s_{i,j}^{x\rightarrow \mathrm{y1M_2}})_{\mathrm{M_2}}).\label{given_1_mask}
\end{equation}
We extract the top-5 words for each of the top-5 $y_1$ candidates, resulting in 25 unique phrases, some of which are single words consisting of two subword tokens. The (normalised) joint probability of $y$: $y_1y_2$ is estimated as:
\begin{gather}
\begin{aligned}
\mathrm{P_2}(y |S_{j}) =\sqrt{\mathrm{\tilde{P}}(y_2|y_1,S_{j})\tilde{\mathrm{P}}(y_{1}|S_{j})}.\label{jointp}
\end{aligned}
\end{gather}
Lastly, we swap the mask-filling order of $y_1$ and $y_2$ and generate another 25 phrases, some of which overlap with the previous 25 phrases. We retain the top-10 paraphrases with the largest joint probabilities and combine them with the top-10 single-token paraphrases.\footnote{We discard candidates that are highly similar to the target MWE (e.g.\ \ex{swan~songs} for \ex{swan~song}) based on a threshold on the normalised character-level edit distance between the candidate and $x$  of 0.2 or less.} While this algorithm could be extended to generating longer phrases, we focus on 1- and 2-token paraphrases based on the observation that many MWEs can be regarded as single semantic units \cite{10.2307/25000002} and are thus paraphrasable with single words (e.g.\ \ex{kick~the~bucket} with \ex{die}) \cite{mwe}. In \secref{MWE_paraphrase}, we also experiment with using T5 \cite{t5} instead of BERT to generate paraprhases without the token-length constraint.

\subsection{Outer Probability Reranking} \label{step4}
After generating paraphrase candidates with differing token lengths, the question is how to jointly rank them. One simple solution is to compare the mask-filling probabilities $\mathrm{P_1}(y|S_j)$ and $\mathrm{P_2}(y|S_j)$ directly. However, these values are not directly comparable because the former tends to get higher values due to the narrower search space. Also, these generation probabilities are affected by the word frequency of $y$, with rare words receiving smaller values.\footnote{For instance, \citet{lau-etal-2020-furiously} measure the acceptability of a sentence based on the LM perplexity and show that it is crucial to normalise the perplexity by unigram probability of the words in the sentence because rare words tend to receive low probabilities even if they are used in a natural context.} As such, we propose a new reranking method based on the ``outer probability'': the probability of generating the {\it context} of $x$ given the candidate. More specifically, for each sentence $s_{i,j}$ in the cluster $j$, we first replace some context words in $s_{i,j}$ with [MASK] tokens to create the masked sentence $\hat{s}_{i,j}$. Then, we replace $x$ with one of the paraphrase candidates $y$ and predict the masked tokens using an MLM. Our hypothesis is that if $y$ represents the semantics of $x$ very well, the model will predict the surrounding words with higher probabilities.\footnote{This resembles the training objective of skip-gram \cite{word2vec}.} Lastly, we calculate the reranking score $\mathrm{S}(y|S_j)$ by taking the average of the log probabilities over all the masked tokens in all the sentences $s_{i,j}$ in cluster $j$:
\begin{gather*}
\mathrm{S}(y|S_j)= \sum_{i = 1}^{N_j}\log \mathrm{P}({s}_{i,j}^{x\rightarrow \mathrm{y}}|\hat{s}_{i,j}^{x\rightarrow \mathrm{y}}),
\end{gather*}
where $\mathrm{P}({s}_{i,j}^{x\rightarrow \mathrm{y}}|\hat{s}_{i,j}^{x\rightarrow \mathrm{y}})$ is the product of the probabilities of reconstructing the masked tokens. One advantage of this method is that the tokens we predict are always the same regardless of $y$, reducing the influence of the number of tokens in $y$. 

Regarding which tokens to mask in ${s}_{i,j}$, one naive approach is to randomly mask words as performed during the MLM pre-training. Ideally, however, we want to select words that are semantically relevant to $x$; for instance, given the context: \ex{This show will serve as his \underline{swan song}, as he plans to retire}, the words \ex{show} and \ex{retire} are arguably more relevant to the meaning of \ex{swan~song} than other words such as \ex{will} and \ex{plans}. With this in mind, we mask words based on the self-attention weights: first, we replace $x$ with two [MASK] tokens and calculate their self-attention weights from the other words in the last layer.\footnote{We average the weights across all attention heads and [MASK] tokens. We also tried using only one [MASK] token instead of two and obtained similar results.} We then mask the top-5 words with the highest weights, excluding punctuation and subword tokens.

\subsection{Paraphrase Retrieval}
Given the ranked paraphrase candidates for each cluster, we finally retrieve the paraphrases of 
$x$ by retrieving the cluster that best represents $x$ in the target sentence $t$. To this end, we first replace $x$ in $t$ with one [MASK] token and retrieve the closest cluster based on the cosine similarity between the [MASK] embedding and the centroids of the clusters. Note that all the previous steps can be done without $t$, meaning if we have a list of potential MWE types and pre-compute their paraphrase candidates, we can paraphrase them given an arbitrary context in an online manner.

\section{Experiments}
\subsection{Idiomatic Semantic Text Similarity Task}

\subsubsection{Data}
We first evaluate our model on SemEval 2022 Task 2 Subtask B \cite{tayyar-madabushi-etal-2022-semeval}. This is a variant of the semantic textual similarity (STS) task, where given two input sentences, a model produces a score between 0.0 and 1.0 based on the similarity of the sentences, and the evalation is based on Spearman's rank correlation to the human-annotated scores. One of the input sentences $E$ contains an MWE (e.g.\ $E$: \ex{Witten's \underline{swan song} was far from a hit}) and the other sentence is a replica of $E$ except that the MWE is replaced with either a correct paraphrase ($E_{\rightarrow c}$, e.g.\ \ex{Witten's \underline{final performance} was far from a hit}) or an incorrect one ($E_{\rightarrow i}$, e.g.\ \ex{Witten's \underline{bird song} was far from a hit}). The target MWEs are two-word nominal compounds that can contain adjectives (e.g.\ \ex{old~flame}). For $E$--$E_{\rightarrow c}$ pairs, the STS score is 1.0 since they have almost identical meaning, while for $E$--$E_{\rightarrow i}$ pairs, sentences are scored in the $[0,1)$ range. Data is provided in English, Portuguese, and Galician, and for English and Portuguese, general STS benchmark data sets --- STS Benchmark \cite{cer-etal-2017-semeval} for English and ASSIN2 STS \cite{assin2} for Portuguese --- are also included in the evaluation set to assess the model's generalisability on both MWE and non-MWE data. 

\subsubsection{Models}

There are two settings for the shared task: Fine-Tune and Pre-Train. In the Fine-Tune setting, systems can be supervised on the train split of the MWE STS data,\footnote{For Galician, there is no train or development data.} but in the Pre-Train setting, systems are not allowed to use this data but can be pre-trained on other resources. Since our model works without any labelled data (or even fine-tuning), we compare our model against  Pre-Train systems. 

In the Pre-Train setting, the best-performing system \cite{phelps-2022-drsphelps} expands the BERT vocabulary and obtains additional embeddings for each MWE type using monolingual data containing the MWEs. As such, it is built on similar data to our model. To obtain the MWE embeddings, they employ BERTRAM \cite{schick-schutze-2020-bertram}, which was originally proposed for learning additional BERT input embeddings of rare words. The model with the MWE embeddings is then fine-tuned on the train split of {\it general} STS data. The major limitation of this approach is that the MWE embeddings need to be pre-trained for each model (e.g.\ BERT-base, BERT-large, T5) before they are fine-tuned. In contrast, our model {\it directly paraphrases the input text} $E$ containing an MWE, and we feed it to an arbitrary pre-trained STS model to generate the sentence embedding of $E$, which is then used to measure the similarity of the sentence pairs.\footnote{The other input text ($E_{\rightarrow c}$ or $E_{\rightarrow i}$) is fed into the STS model without any paraphrasing.} That is, our paraphrasing model is completely separated from the task-specific models, providing more flexibility in terms of model selection and training. In the STS experiments, however, we use the same BERT models used in \citet{phelps-2022-drsphelps} for both paraphrasing and STS models (and also for clustering) for fair comparison. Specifically, we use BERT-base-uncased \cite{bert} for English, BERTimbau-Base \cite{souza2020bertimbau} for Portuguese, and Bertinho-Base \cite{bertinho} for Galician. To obtain STS models, we fine-tune the respective BERT models on the train split of STS Benchmark for English and ASSIN2 STS for Portuguese and Galician, following \citet{phelps-2022-drsphelps}.\footnote{The Portuguese data is used for Galician since there is no STS Galician data set and these languages are very similar.} 

\begin{table}[t!]
\begin{center}
\begin{adjustbox}{max width=\columnwidth}
\begin{tabular}{lccc@{\;}ccc@{\;}ccc@{\;}ccc@{\;}}
\toprule
Model&All&MWE&General\\\midrule
\multicolumn{4}{c}{Unsupervised}\\\midrule
\multirow{1}{*}{Sem-Base} &
 48.10&22.63&83.11\\

\multirow{1}{*}{\citet{phelps-2022-drsphelps}} &
 64.02&40.30&86.41\\

\multirow{1}{*}{\textbf{OURS}} &
65.31&42.65&\textbf{86.91}\\
\multirow{1}{*}{\textbf{OURS-ave3}} &
\textbf{66.13}&\textbf{42.68}&\textbf{86.91}\\

\midrule
\multicolumn{4}{c}{Supervised}\\\midrule

\multirow{1}{*}{Sem-Base} &
 59.51&39.90&59.61\\
\multirow{1}{*}{\citet{phelps-2022-drsphelps}} &
65.04&41.24 &81.88 \\
\multirow{1}{*}{\citet{liu-etal-2022-ynu}} &
  66.48&42.77&66.37&\\

\bottomrule
\end{tabular}
\end{adjustbox}
\end{center}
\caption{Results (Spearman's rank correlation $\times100$) on the SemEval STS task. The best scores among the unsupervised models are boldfaced.}
\label{result_sts}
\end{table}

\begin{table}[t!]
\begin{center}
\begin{adjustbox}{max width=\columnwidth}
\begin{tabular}{cc@{\;\;}ccc}
\toprule
 &Model&All&MWE&General\\\midrule
\multirow{3}{*}{EN} &\citet{phelps-2022-drsphelps}
 &74.45& 44.22& 87.09\\
&+ Fine-Tune&\textbf{76.43}&48.61&83.44\\\cmidrule(lr){2-5} 
&{OURS-ave3}&76.31&\textbf{50.43}& \textbf{88.74}\\\midrule
\multirow{3}{*}{PT} &\citet{phelps-2022-drsphelps}
 &70.87& \textbf{48.06}& 80.10\\
 &+ Fine-Tune&73.07&46.43& 79.08\\  \cmidrule(lr){2-5}

&{OURS-ave3}&\textbf{73.97}& 45.30& \textbf{80.54}
\\\midrule
\multirow{3}{*}{GL} &\citet{phelps-2022-drsphelps}
 &---&{29.24} &---\\
 &+ Fine-Tune&---&28.59&---\\\cmidrule(lr){2-5}
&{OURS-ave3}&---& \textbf{34.74}&---\\

\bottomrule
\end{tabular}
\end{adjustbox}
\end{center}
\caption{Results on the SemEval STS task for each language. The best scores for each language are boldfaced.}
\label{result_sts_lang}
\end{table}
\subsubsection{Results}
\tabref{result_sts} shows the results on the STS task, where the columns ``MWE'' and ``General'' denote the Spearman's rank correlation on the MWE and general STS data, and ``All'' denotes the overall performance. \textbf{Sem-Base} denotes the mBERT-based baseline provided by the shared task organisers.  \textbf{OURS} and \textbf{OURS-ave3} denote our models, where the latter indicates the performance when we produce the sentence embedding of $E$ by replacing the target MWE with the top-3 paraphrases and averaging the embedddings of the three sentences.  In the shared task, the systems are ranked based on the ``All'' performance, and the table shows that our model outperforms \citet{phelps-2022-drsphelps} and achieves the best score in the Pre-Train setting. It also shows that using the top-3 paraphrases further improves performance. 

The ``Supervised'' sub-table shows the performance of the submitted systems in the Fine-Tune setting, where the models are trained on both MWE and general STS data. Without any labelled data, our model outperforms the supervised task baseline (Sem-Base) and also the supervised model of \citet{phelps-2022-drsphelps}, which fine-tunes their unsupervised model on the MWE STS data and ranks 2nd in the Fine-Tune setting. OURS-ave3 performs slightly worse than the best supervised system \cite{liu-etal-2022-ynu}, which however performs very poorly on the general STS data, suggesting it is overfitting the MWE data. \tabref{result_sts_lang} compares the performance of \citet{phelps-2022-drsphelps} (w/ or w/o fine-tuning) and our model for each language. Overall, our model achieves better performance than the unsupervised baseline in all languages, and the supervised model in Portuguese and Galician. We expect further improvements by fine-tuning BERT on labelled or unlabelled data, which we leave for future work. 

\subsection{MWE Paraphrase}\label{MWE_paraphrase}

Next, we evaluate our model based on the matching accuracy of the MWE paraphrases. To this end, we first extract the pairs of MWEs and their correct paraphrases in context from the English ``train'' split of the SemEval STS data set provided for the Fine-Tune setting. Since we do not use any portion of this data to train or tune our model, we can regard it as a pseudo test set for this task;\footnote{We could not use the test set because the gold paraphrases are not publicly available.} the size of each data split is shown in \tabref{N_MWE} in Appendix. In this task, in addition to BERT-base, we also combine our method with BERT-large (whole word masking), SpanBERT-large \cite{spanbert}, ALBERT-large \cite{albert} and T5 \cite{t5}.\footnote{See the Appendix for model details (\tabref{hyper_param_table}).} Unlike BERT and its variants, T5 generates an arbitrary number of tokens conditioned on the encoder hidden states of all input tokens, making it difficult to aggregate the hidden states or probability distributions across multiple sentences as done in \eqnref{bert_1mask}. Therefore, we propose another simple method tailored for T5 (which is somewhat different from the method described in \secref{step3}): we use it to generate 20 paraphrase candidates for each sentence $s_{i,j}$ in the cluster $j$\footnote{We use the T5 encoder to create the clusters.} independently using beam search, which produces $20N_j$ paraphrases in total.\footnote{To avoid generating lengthy phrases, we set the maximum number of tokens to generate to 10.} Then, we retain the paraphrases that contain 1 or 2 {\it words} (which can consist of more subword tokens)\footnote{We also tried including 3-word expressions but got worse results as most gold paraphrases are 1 or 2 words.} and rank them based on their generated counts; for the 2-word candidates, we double the count because one-word candidates often get higher values, leading to worse results. We also try reranking the candidates using the outer probability (\secref{step4}), but randomly masking 5 consecutive words for T5 (and SpanBERT) rather than the high-attention  words because these models are trained to fill random {\it spans} of text rather than separate tokens. As a strong baseline for this task, we use GPT-3 ({\it davinci}) \cite {gpt-3}. As a prompt, we feed several triples randomly retrieved from the dev set, which consist of a sentence that contains an MWE, a question asking what is the most appropriate substitute for it, and the correct paraphrase. We feed as many examples as possible until they reach the max token limit (2048), which correspond to about 35 triplets. \citet{swords} show that this approach outperforms BERT in lexical substitution, and we use their prompt with minor modification.\footnote{See Appendix \ref{gpt3_prompt} for an example of the prompt.}

\begin{table}[t!]
\begin{center}
\begin{adjustbox}{max width=\columnwidth}
\begin{tabular}{lc@{\;\;}ccccc}
\toprule
 \multirow{2}{*}{Model}&\multirow{2}{*}{\# param}&\multicolumn{3}{c}{Matching Accuracy}&&STS\\
 \cmidrule{3-5}\cmidrule{7-7}
 &&P@1&P@5&P@10&&$\rho$\\\midrule
GPT-3&175B                    &\textbf{13.2}& ---& ---&&74.2\\\midrule
BERT-base&110M                & \z8.2& {18.4}& {24.3}&&76.3\\
BERT-large&340M               & \z8.1& 19.4& {27.6}&&76.1\\
SpanBERT&340M                 & \z8.5&22.0&{28.9}&&76.2\\
ALBERT&\z17M                  & \z6.6& 13.5& 21.3&&75.4\\
T5-base&\multirow{2}{*}{220M} &\textbf{10.8}&22.0&25.8&&73.6\\
~+ rerank&                    &10.1&22.6&27.1&&\textbf{76.9}\\
T5-large&\multirow{2}{*}{770M}&\z8.0&22.5&28.0&&74.4\\
~+ rerank&                    &\z7.2&\textbf{25.6}&\textbf{33.3}&&76.0\\

\bottomrule
\end{tabular}
\end{adjustbox}
\end{center}
\caption{The performance (\ensuremath{\text{P@}k}) of GPT-3 (baseline) and our models on English MWE paraphrasing and STS tasks (``\# param'' = model parameter size). The best scores are boldfaced.}
\label{result_paraphrase}
\end{table}

\tabref{result_paraphrase} (under ``Matching Accuracy'') shows the results based on \ensuremath{\text{P@}k}: the proportion of instances where the gold paraphrase is included in the top-$k$ predictions. For GPT-3, we report only \ensuremath{\text{P@}1} because it is prompted to produce the single best paraphrase only, as we have only one gold paraphrase per input sentence. We can see that GPT-3 performs the best in P@1, followed by our models using T5 and BERT.\footnote{One reason why P@1 is very low is that the SemEval data contains only one gold paraphrase for each MWE (e.g.\ \ex{final performance} for \ex{swan song}), missing many other possible candidates (e.g.\ \ex{last performance}, \ex{final appearance}, and \ex{farewell appearance}).} However, note that they are not strictly comparable in terms of the number of the model parameters (shown as ``\# param'')  as well as the amount of training data (with BERT trained on the least data). Also, GPT-3 is supervised on labelled data while our models are fully unsupervised.\footnote{Also, GPT-3 implicitly memorises the definitions of some MWEs during pre-training, e.g.\ given a simple prompt like \ex{What is the meaning of ``swan song''?}, it returns its origin as well as its meaning.} In \ensuremath{\text{P@}5/10}, T5-large performs the best, but using much more parameters than the other models. Our reranking method improves the performance of T5 overall, demonstrating its effectiveness. BERT performs reasonably well at \ensuremath{\text{P@}10} with fewer parameters, possibly because BERT is conditioned on all sentences $s_{i,j}$ in the cluster $j$ simultaneously by averaging the mask embeddings.

We also evaluate the models on the STS task (using the English test split), and the result is shown under ``STS'' in \tabref{result_paraphrase}. Notably, GPT-3 performs the worst in this metric, and this is likely because it sometimes copies the target MWEs or paraphrases them with another MWE (e.g.\ \ex{dead end} with \ex{blind alley}), which does not simplify the text very much. The comparison of the other models also reveals that larger models are not necessarily better at text simplification.

\begin{table}[t!]
\begin{center}
\begin{adjustbox}{max width=\columnwidth}
\begin{tabular}{lccccccccc}
\toprule
\multirow{2}{*}{{Method}}& \multicolumn{3}{c}{\ensuremath{\text{P@}k} (EN)}&&\multicolumn{3}{c}{\ensuremath{\text{P@}k} (PT)}\\\cmidrule{2-4}\cmidrule{6-8}
&\ensuremath{1}&\ensuremath{5}&\ensuremath{10}&&\ensuremath{1}&\ensuremath{5} &\ensuremath{10}\\\midrule
\multirow{1}{*}{{None}} &6.5&
15.8&\textbf{24.4}&&9.7&15.9&\textbf{21.1}\\
\multirow{1}{*}{{$K$-means (2)}} &7.7&
16.5&23.2&&9.8&16.4&18.5\\
\multirow{1}{*}{{$K$-means (3)}} &7.2&
16.0&22.5&&9.9&15.9&18.4\\
\multirow{1}{*}{{$K$-means (4)}} &6.2&
13.8&21.0&&10.1&15.2&16.5\\
\multirow{1}{*}{{X-means}} 
&7.8&
16.4&24.3&&9.5&16.9&19.0\\
\multirow{1}{*}{{DBSCAN}} 
&\textbf{8.2}&\textbf{18.4}&24.3&&\textbf{10.6}&\textbf{18.3}&20.9\\
\bottomrule
\end{tabular}
\end{adjustbox}
\end{center}
\caption{The performance of our approach with different clustering methods on the MWE paraphrasing task.}
\label{analysis_clustering_paraphrase}
\end{table}

\begin{table*}[t!]
\begin{center}
\begin{adjustbox}{max width=\textwidth}
\begin{tabular}{cl}
\toprule
\multirow{1}{*}{Top-1 paraphrase}& Sampled sentences from each cluster\\\midrule

\makecell{--} &There is no magic pill or \textbf{silver bullet} because each of us is different.\\
\makecell{perfect solution}&But it isn’t a \textbf{silver bullet} and drawbacks exist. \\
\makecell{single practical}&There is no \textbf{silver bullet} machine learning algorithm that works well across all problem spaces.\\ 
\midrule

\makecell{--} &You don't even need to be \textbf{closed book} as you do the practice exam questions, if you don't feel ready to.\\

\makecell{book} &Following my prayer, I held the \textbf{closed book} in my hands and turned to today’s passage in the day book.\\

\makecell{complete mystery} &The answer to this question is still a \textbf{closed book} for some modern historians.\\
\makecell{fully written}&Complete the \textbf{closed book} exam without removing questions or answers.\\\midrule

\makecell{--} &Does anyone in the \textbf{inner circle} know if he's ok?\\
\makecell{personal staff} &What good leaders do is recognize talent and choose their \textbf{inner circle} and cabinet based on that knowledge.\\
\makecell{small group} &I am very picky about who I let into my \textbf{inner circle} of friends for this very reason.\\
\makecell{array} &  Developing an \textbf{inner circle} of great thinkers within the organization you consider your partners.\\
\midrule
\makecell{--} &This may seem like a \textbf{small fry} but it isn’t.\\
\makecell{nothing special} &It just seems like \textbf{small fry} compared to the struggles of last century.\\
\makecell{young person} &and i'm just a \textbf{small fry} in society.\\
\makecell{small metal} & Add it to a \textbf{small fry} pan with some cooking spray to heat and brown it a bit.\\

\bottomrule
\end{tabular}
\end{adjustbox}
\end{center}
\caption{The top-1 predicted paraphrases and example sentences (with MWEs shown in bold font) for each cluster generated by our model using DBSCAN. The first sentence for each MWE is sampled from the outlier cluster, which is discarded and not used for paraphrase generation (and thus no paraphrase is given).}
\label{examples_cluster}
\end{table*}

\section{Analysis}

\subsection{Effects of Clustering}\label{effects_clustering_sec}
We compare the performance of our model (BERT-base) using different clustering methods: $K$-means, X-means and DBSCAN. \tabref{analysis_clustering_paraphrase} shows the results on the MWE paraphrasing task. The first row ``None'' denotes the performance when we treat all sentences as a single cluster. Amongst all methods, DBSCAN performs the best at \ensuremath{\text{P@}1/5}, outperforming $K$-means and X-means. We conjecture that DBSCAN benefits from not only setting the cluster size per MWE type dynamically, but also from creating one outlier cluster and discarding sentences that do not provide sufficient context to infer the MWE semantics (which is not achieved by $K$/X-means). Without clustering, our method performs poorly at \ensuremath{\text{P@}1/5}, and yet is equivalent to DBSCAN at \ensuremath{\text{P@}10}. This is because our model with clustering completely fails when it retrieves a wrong cluster,\footnote{This is particularly pronounced in the low performance of $K$-means ($K=4$) in \ensuremath{\text{P@}10}.} whereas our model without clustering always generates a mixed bag of paraphrases with varying meanings, some of which are often relevant to the (common) MWE senses.

\tabref{examples_cluster} shows some examples of sentences\footnote{The input text fed to the clustering model usually contains the previous and/or following sentences as well.} sampled from each DBSCAN cluster, as well as the best paraphrase generated for each cluster. The first sentences for each MWE are ``outlier sentences'' sampled from the outlier cluster, and are thus not used for paraphrase generation. We can see that the outlier sentences of \ex{inner~circle} and \ex{small~fry} have vague contexts that do not represent the MWE semantics very well. In contrast, the one of \ex{silver~bullet} is regarded as an outlier despite its specific context because the target MWE is conjoined with another MWE (\ex{magic~pill}) and the model wrongly predicts \ex{pill} and \ex{drag} as substitutes for \ex{silver~bullet}.

Comparing the three clusters of \ex{closed~book} and their paraphrases, we can see that the paraphrases generated by the first and second clusters successfully distinguish its literal and idiomatic senses (\gl{book} and \gl{complete mystery}), but the paraphrase of the last cluster (\ex{fully written}) does not retain the original meaning; in fact, it is very challenging to find a good (short) substitute for \ex{closed~book} in this context, as an inherent limitation of a paraphrasing approach. The table also shows that some clusters are formed based on {\it syntactic} constraints rather than the semantics of the MWEs; e.g.\ the clusters of \ex{silver~bullet} are formed based on whether the MWE is used as a noun or adjective. The last cluster of \ex{inner~circle}, which has similar semantics to the second cluster, is also created as a result of the local morphophonetic effect, where most of the sentences in this cluster include the MWE as \ex{\textbf{an}~\underline{inner circle}~of} and {\it all} the paraphrase candidates start with a vowel sound (e.g.\ \ex{entire army}, \ex{elite group}, \ex{alliance}) due to the article--noun agreement effect. A similar result is observed by \citet{wada-etal-2022-unsupervised} in lexical substitution, where they find that MLM predictions and representations are highly affected by such morphophonetic or morphosyntactic biases. Interestingly, the last cluster of \ex{small~fry} is composed of the sentences where \ex{fry} is used as a component word of the other MWE  \ex{fry~pan}, not of \ex{small~fry}. Other similar cases include \ex{high life} (\gl{expensive lifestyle}) used as \ex{high life expectancy} and \ex{bad hat} (\gl{troublemaker}) used as \ex{bad hat hair}. We further discuss this problem in \secref{limitation}.

\begin{table}[t!]
\begin{center}
\begin{adjustbox}{max width=\columnwidth}
\begin{tabular}{cccccccccc}
\toprule
\multirow{2}{*}{{Model}}&\multirow{2}{*}{{Method}}& \multicolumn{3}{c}{\ensuremath{\text{P@}k} (EN)}&&\multicolumn{3}{c}{\ensuremath{\text{P@}k} (PT)}\\\cmidrule{3-5}\cmidrule{7-9}
&&\ensuremath{1}&\ensuremath{5}&\ensuremath{10}&&\ensuremath{1}&\ensuremath{5} &\ensuremath{10}\\\midrule
\multirow{3}{*}{{BERT-b}}&\multirow{1}{*}{{None}} &\textbf{9.6}&
16.8&21.2&&7.6&13.5&19.2\\
&\multirow{1}{*}{{Rand}} &6.9&
17.1&\textbf{25.2}&&\textbf{10.8}&17.6&20.0\\
&\multirow{1}{*}{{Attn}} 
&{8.2}&\textbf{18.4}&24.3&&{10.6}&\textbf{18.3}&\textbf{20.9}\\
\midrule
\multirow{3}{*}{{BERT-l}}&\multirow{1}{*}{{None}} &\textbf{11.6}&
21.3&26.3&&10.1&17.7&\textbf{25.9}\\
&\multirow{1}{*}{{Rand}} &6.9&
\textbf{21.8}&\textbf{29.3}&&11.0&21.1&23.6\\
&\multirow{1}{*}{{Attn}} 
&8.1&{19.4}
&{27.6}&&\textbf{11.3}&\textbf{21.6}&{23.8}\\
\bottomrule
\end{tabular}
\end{adjustbox}
\end{center}
\caption{The MWE paraphrasing performance of our approach using BERT-base/large and different reranking methods. ``Rand'' is the average score over 3 runs.}
\label{analysis_reranking}
\end{table}
\begin{table*}[t]
\begin{center}
\begin{adjustbox}{}
\begin{tabular}{cccc@{\;}ccc@{\;}ccc@{\;}ccc@{\;}}
\toprule
\multirow{2}{*}{{Method}} &\multicolumn{2}{c}{ALL}&&\multicolumn{2}{c}{EN}&&\multicolumn{2}{c}{ PT}&&\multicolumn{1}{c}{GL}\\\cmidrule{2-3}\cmidrule{5-6}\cmidrule{8-9}\cmidrule{11-11}
& All&MWE&&All&MWE&&All&MWE&&MWE\\\midrule

\multicolumn{13}{c}{dev} \\\midrule

\multirow{1}{*}{{None}} 
&80.61&39.41&&86.26&49.99&&78.07&37.83&&---\\

\multirow{1}{*}{{Rand}} &80.14&37.65&&86.08&48.97&&77.18&34.80&&---\\

\multirow{1}{*}{{Attn}} &{\textbf{81.24}}&{\textbf{41.75}}&&{\textbf{86.70}}&{\textbf{51.86}}&&\textbf{78.31}&\textbf{38.28}&&---\\\midrule

\multicolumn{13}{c}{test} \\\midrule

\multirow{1}{*}{{None}} &65.39&41.48&&75.61&49.21&&73.76&44.30&&33.98\\

\multirow{1}{*}{{Rand}} &65.41&41.42&&75.93&49.49&&73.93&\textbf{45.44}&&32.19\\
{{Attn}} &\textbf{66.13}&\textbf{42.68}&&\textbf{76.31}&\textbf{50.43}&&\textbf{73.97}&45.30&&\textbf{34.74}\\
\bottomrule
\end{tabular}
\end{adjustbox}
\end{center}
\caption{The STS performance (Spearman's rank correlation$\times100$) of our models (OURS-ave3) using different reranking methods. The scores of Rand are averaged over 3 runs. The best scores in each data split are shown in bold. The general STS scores are omitted from the table as they stay the same across all models (86.93/92.79 on the dev set and 88.74/80.54 on the test set for EN/PT).}\label{analysis_rerank_all}
\end{table*}

\subsection{Effects of Reranking}

In \tabref{result_paraphrase}, we showed that both T5-base and T5-large benefit from reranking. To further verify its effectiveness, we compare the performance of BERT-base/large w/ or w/o reranking in \tabref{analysis_reranking}. The row ``None'' denotes the performance when we rank the candidates based on the mask-filling probabilities $\mathrm{P_1}(y|S_j)$ and $\mathrm{P_2}(y|S_{j})$ in \eqnref{bert_1mask} and \eqnref{jointp}, and ``Rand'' and ``Attn'' show the performance with our reranking method, wherein we mask five words randomly or based on self-attention weights; for Rand, we average the performance over three runs. Our reranking method improves the \ensuremath{\text{P@}5/10} performance of BERT-base in both English and Portuguese, but for BERT-large, there is no clear winner. When we compare Rand and Attn, there is no noticeable difference in matching accuracy; therefore, we also examine the impact of our reranking methods on the STS task in \tabref{analysis_rerank_all} (which can measure the semantic fit of the paraphrases in a continuous manner). It shows that Attn performs the best overall on the dev and test sets. We also observe that the performance of Rand varies greatly across three runs (65.34, 65.91, and 64.96 in ALL on the test set), suggesting that the selection of the context words to reconstruct in our reranking method has a non-trivial impact on performance.

\tabref{rerank_example} shows examples of MWEs and their top-3 paraphrases generated by BERT-base w/ or w/o reranking (Attn). It shows that reranking produces semantically more relevant paraphrases. Furthermore, these results reveal a few plausible limitations of the scoring method based on the mask-filling probabilities; the limitations that are also relevant to T5 to some degree. Firstly, it tends to favour single-token paraphrases (e.g.\ \ex{bridge} and \ex{road} for \ex{zebra crossing}) due to the narrower search space. Secondly, it often assigns 
high scores to phrases that have strong collocation (e.g.\ \ex{new world}, \ex{regular customer}) regardless of their semantic fit because of the higher probability $\mathrm{\tilde{P}}(y_2|y_1,S_j)$ in \eqnref{given_1_mask} (or $\mathrm{\tilde{P}}(y_1|y_2,S_j)$). Thirdly, it is inevitably influenced by word frequency, as evidenced by the fact that \ex{research \underline{organization}} gets much higher values than its British spelling \ex{research \underline{organisation}} (0.33 vs.\ 0.09) as the paraphrase of \ex{think~tank}; on the other hand, our reranking method assigns them very similar scores. On the other hand, one limitation of our reranking method is that it occasionally gives high scores to syntactically ill-formed candidates, especially those that contain duplicated tokens (e.g.\ \ex{clock~clock} as the paraphrase for \ex{grandfather~clock}), and we alleviate this by removing one of the duplicated tokens.

\begin{table}[t!]
\begin{center}
\begin{adjustbox}{max width=\columnwidth}
\begin{tabular}{cccc@{\;}ccc@{\;}ccc@{\;}ccc@{\;}}
\toprule
MWE&w.o. Rerank&w. Rerank\\\midrule

zebra crossing&\makecell{bridge  \\ pedestrian bridge \\road}&\makecell{\textbf{pedestrian crossing}\\pedestrian bridge\\road crossing} \\  \midrule 
melting pot&\makecell{new world  \\ unique mix \\diverse mix}&\makecell{\textbf{mixture}\\unique mix \\collection} \\ \midrule  

think tank&\makecell{research organization\\organization\\\textbf{research group}}  &\makecell{research organization \\research organisation \\\textbf{research group}}\\\midrule
busy bee&\makecell{\textbf{busy person}\\good person\\regular customer}  &\makecell{\textbf{busy person}  \\busy woman \\busy man }\\

\bottomrule
\end{tabular}
\end{adjustbox}
\end{center}
\caption{Examples of the top-3 paraphrases predicted by our model w/ or w/o reranking (Attn). The gold paraphrases are boldfaced.}
\label{rerank_example}
\end{table}

\section{Related Work}
There is a line of work on MWE paraphrasing (or substitution), but unlike our method, previous methods resort to either human-annotated corpora or high-coverage dictionaries to generate paraphrases; consequently, they are only evaluated in English or other high-resource languages (e.g.\ Chinese). For instance, \citet{liu-hwa-2016-phrasal} extract paraphrases of MWEs from their dictionary definitions, which usually contain supplementary information as well as their core meanings and paraphrases. Similarly, \citet{Zhou_Zeng_Gong_Bhat_2022} encode definitions of MWEs using a sentence embedding model and employ the embeddings to generate their paraphrases. They also propose another method that fine-tunes BART \cite{bart} on a parallel corpus built by \citet{zhou-etal-2021-pie}, in which the source sentences contain MWEs and the target sentences paraphrase them while leaving the other words unchanged. Similarly, \citet{chinese_para} create analogous data in Chinese and fine-tune mT5 \cite{mt5} on it to paraphrase Chinese MWEs. In contrast, \citet{ponkiya2020} 
propose an unsupervised method using BERT/T5, but they focus on paraphrasing noun compounds (e.g.\ \ex{club house}) using the same MWE tokens (e.g.\ \ex{\underline{house} owned by a \underline{club}}) without context.

A more common and resource-efficient approach to handling MWEs is to regard them as individual lexical units (e.g.\ regard ``kick\_the\_bucket'' as one token) and train their embeddings using monolingual data \cite{salehi-etal-2015-word, 10.1162/coli_a_00341,phelps-2022-drsphelps}. Those embeddings can be used to retrieve similar words in the vocabulary based on embedding similarity \cite{pre_tok_bli}. However, one drawback is that it increases the size of vocabulary 
(and parameters) significantly, due to the sheer volume of MWE instances (e.g.\ around 41\% of entries in WordNet 1.7 \cite{wordnet} are MWEs \citep{Pain_in_the_neck}).  Recently, \citet{bart_idiom} addressed this limitation by training an additional adapter network \cite{adapterhub} on top of an MLM to produce better embeddings for various MWEs, but they rely on dictionary definitions to train the network, arguing that such external knowledge is fundamental for learning MWE representations.

\section{Conclusion}
We propose a fully unsupervised method to paraphrase multiword expressions (MWEs) in context. Our method employs only a monolingual corpus and pre-trained language model, and does not rely on any labelled data. In our experiments, we show that our model generates good MWE paraphrases and aids pre-trained sentence embedding models to represent sentences containing MWEs.

\section{Limitations}\label{limitation}
One limitation of our proposed method is that it requires the pre-identification of the target MWE in a sentence before paraphrasing it, a task that is not a walk in the park. In particular, it is very challenging to identify what is the ``correct'' span of a given MWE, which our model critically relies on. For instance, given the MWE \ex{lip~service} (\gl{insincere agreement}), our model predicts \ex{more attention} as the best paraphrase, likely because the MWE is usually used as \ex{pay~lip~service~to~(something)}, and \ex{attention} is one of the few words that fits well in this context (in terms of collocation). Therefore, the whole phrase \ex{pay~lip~service~to} should be identified as an MWE instance\footnote{In fact, it is registered as such in some English dictionaries.} when it is used in sentences like \ex{They \underline{pay lip service to} the idea}; however, \ex{lip~service} can also serve as one lexical unit in sentences like \ex{It wasn't just \underline{lip service}}. A similar problem arises when we deal with nominal MWEs that follow indefinite articles (\ex{a} or \ex{an}) as discussed in \secref[]{effects_clustering_sec}, or verbal MWEs that are often followed by specific prepositions (e.g.\ \ex{turn~a~blind~eye~\textbf{to} ...} means \gl{deliberately ignore ...}) because the MLM prediction is affected by the syntactic constraint.\footnote{In languages where words have grammatical gender such as Portuguese and Italian, this problem can be more pronounced because context words including adjectives and determiners are affected by gender.} MWE span identification is also important in our sentence collection process; e.g.\ as discussed in \secref{effects_clustering_sec}, the phrase \ex{small~fry} can be used as \ex{small {fry pan}} rather than as the MWE meaning \gl{insignificant}, and hence collecting sentences based on string match resulted in one additional cluster that is not relevant to either its literal or idiomatic senses. 

Another limitation is that our model cannot handle discontinuous MWEs such as \ex{\underline{throw} someone \underline{under the bus}} and \ex{\underline{not} ... \underline{in the least}} because it is not clear which parts to mask and paraphrase in such cases. Similar problems arise when continuous MWEs undergo either internal modification (e.g.\ \ex{\underline{go} completely \underline{cold turkey}}) or drastic syntactic transformation (e.g.\ \ex{the beans are split}). However, note that all of these types of expressions, as well as the pre-tokenisation problem discussed above, become a pain in the neck for any approach that regards an MWE  as a lexical unit and learns its holistic embedding.

Lastly, our method heavily relies on the quality of the clusters and is thus prone to error propagation. For instance, our model using BERT always generates \ex{large fish} as the best paraphrase for the MWE \ex{big~fish} and fails to capture its idiomatic sense (\gl{an important person}), likely due to its rare occurrence in monolingual corpora (compared to its literal sense). One possible solution to this problem is to derive more senses by allowing the clustering method to create more clusters with fewer instances, but that institutes a trade-off between accommodating rare senses and creating too many clusters for common senses; hence, there is no silver bullet. In fact, this problem pertains to the longstanding question (with no single correct answer) among lexicographers: how to ``split'' and ``lump'' senses of words, and how fine-grained the sense distinctions should be \cite{hanks_2000,hanks_2012}. 

\bibliography{tacl2021}
\bibliographystyle{acl_natbib}
\appendix

\section{Hyper-Parameters}\label{hyper_param}
\tabref{hyper_param_table} shows the hyper-parameters of DBSCAN: \minPts: the minimum number of points required to form a core point; and $\epsilon$: the maximum distance between two points to be considered as neighbours. We tune $\epsilon$ for each model, and the table shows their shortcut names in the Transformers library \cite{wolf-etal-2020-transformers}. For SpanBERT, we used the model in the original GitHub repository (\url{https://github.com/facebookresearch/SpanBERT}).

\begin{table*}[t]
\begin{center}
\begin{adjustbox}{max width=\textwidth}
\begin{tabular}{ccc}
\toprule
\multirow{1}{*}{Parameters}  &Model&Value\\\midrule
\multirow{1}{*}{{\minPts}} &---& max$(3, \round{0.03N})$\\\midrule
    \multirow{9}{*}{$\epsilon$} &bert-base-uncased&0.4\\
    &bert-large-uncased-whole-word-masking&0.5\\
    &spanbert-large-cased&0.3\\    
    &albert-large-v2&0.3\\    
    &google/t5-v1\_1-base&0.4\\    
    &google/t5-v1\_1-large&0.4\\    
    &neuralmind/bert-base-portuguese-cased&0.3\\
    &neuralmind/bert-large-portuguese-cased&0.3\\
        &dvilares/bertinho-gl-base-cased&0.3\\

\bottomrule
\end{tabular}
\end{adjustbox}
\end{center}
\caption{The hyper-parameters of DBSCAN for each model tuned on the STS dev set. $N$ denotes the number of the sampled sentences.}
\label{hyper_param_table}
\end{table*}




\section{GPT-3 Prompt}\label{gpt3_prompt}
Here is an example of the GPT-3 prompt used in \secref{MWE_paraphrase}:
\begin{quote}  
\tt
Witten's **swan song** was far from a hit.

Q: What is the most appropriate substitute for **swan song** in the above text?

A: final performance
\end{quote}
It is a triple of the target sentence with the target MWE marked with \ex{**}, a question that asks the most approproate substitute for the MWE, and the gold paraphrase retrieved from the dev set. We borrow this template from \citet{swords}, but change \ex{What are appropriate substitutes} to \ex{What is the most appropriate substitute} since we have only one gold paraphrase for each input text. We feed as many triples as possible until they reach the max token limit (2048), which correspond to about 35 triplets. Then, we append one ``test triplet'' that contains one sentence from the test set, the corresponding question, and the answer without the gold paraphrase (i.e.\ \ex{A:}), and make the model predict the paraphrase. We make sure that the MWE in the test triplet is not included in the triplets retrieved from the dev set.

\begin{table}[t!]
\begin{center}
\begin{adjustbox}{max width=\columnwidth}
\begin{tabular}{lcccccccc}
\toprule
Split&\multicolumn{3}{c}{\# MWE types}&&\multicolumn{3}{c}{\# Sents with MWEs}\\ \cmidrule{2-4}\cmidrule{6-8}
&EN&PT&GL&&EN&PT&GL\\\midrule

train &214&108&0&&4,725&1,847&0\\
dev &30&20&0&&521&454&0\\
test  &50&50&50&&1,419&1,124&1,367\\
\bottomrule
\end{tabular}
\end{adjustbox}
\end{center}
\caption{The numbers of the MWE types and target sentences that contain them in each STS data set. The train split is used as a pseudo test set for the MWE paraphrasing task in \secref[]{MWE_paraphrase}.}\label{N_MWE}
\end{table}

\end{document}